\pdfoutput=1

\documentclass[11pt]{article}
\usepackage{coling2020}
\usepackage{times}
\usepackage{latexsym}

\usepackage[utf8]{inputenc}
\usepackage{subcaption}
\usepackage{graphicx}
\usepackage{booktabs}
\usepackage{float}
\usepackage{pgfplots}
\usepackage{tikz}

\usepackage{amsmath}
\usepackage{ifthen}
\usepackage{makecell}
\usepackage{multirow}
\usepackage{adjustbox}
\usepackage{xcolor}
\usepackage{enumitem}
\usepackage{amssymb}
\usepackage{url}

\pgfplotsset{compat=1.15}

\definecolor{lightpurple}{RGB}{185, 139, 232}
\definecolor{lightpink}{RGB}{252, 113, 196}
\definecolor{darkcyan}{RGB}{66, 215, 244}

\definecolor{myorange}{rgb}{1,0.5,0}
\definecolor{darkgreen}{RGB}{110, 160, 88}
\definecolor{myred}{RGB}{160, 70, 49}
\definecolor{mypink}{RGB}{253, 170, 255}
\definecolor{mypurple}{RGB}{120, 33, 122}
\definecolor{darkyellow}{RGB}{176, 188, 83}
\definecolor{darkgrey}{RGB}{160, 160, 155}
\definecolor{brightgreen}{RGB}{131, 191, 47}

\usepgfplotslibrary{fillbetween}
\usepgfplotslibrary{statistics}

\colingfinalcopy 

\title{Exploring the Value of Personalized Word Embeddings}

\author{Charles Welch, Jonathan K. Kummerfeld, Verónica Pérez-Rosas \and Rada Mihalcea  \\
Computer Science \& Engineering \\
University of Michigan \\
\texttt{\{cfwelch,jkummerf,vrncapr,mihalcea\}@umich.edu}
}

\date{}

\begin{document}
\maketitle
\begin{abstract}
In this paper, we introduce personalized word embeddings, and examine their value for language modeling. We compare the performance of our proposed prediction model when using personalized versus generic word representations, and study how these representations can be leveraged for improved performance. We provide insight into what types of words can be more accurately predicted when building personalized models. Our results show that a subset of words belonging to specific psycholinguistic categories tend to vary more in their representations across users and that combining generic and personalized word embeddings yields the best performance, with a 4.7\% relative reduction in perplexity. Additionally, we show that a language model using personalized word embeddings can be effectively used for authorship attribution.
\end{abstract}

\section{Introduction}

\blfootnote{
    %
    %
    \hspace{-0.65cm}  
    This work is licensed under a Creative Commons 
    Attribution 4.0 International License.
    License details:
    \url{http://creativecommons.org/licenses/by/4.0/}.
}

Word embeddings have become ubiquitous in natural language processing applications. Usually, embeddings are trained from a large corpus of news or web data that contains writing from many sources and authors~\cite{mikolov2013efficient,pennington-etal-2014-glove}. These embeddings capture syntactic and semantic properties of the language of all authors who contributed to this corpus.

Multi-source corpora provide large volumes of data, but they may not lead to the ideal representations for individuals.
For instance, the word ``hometown'' may have a different  representation for different individuals. For some, it may relate to words such as ``hills,'' ``trees,'' and ``family,'' whereas for others may be more strongly connected to ``ocean,'' ``beach,'' and ``friends.''  These personalized representations differ among individuals, and also differ from a more generic representation that often tends to capture words that are semantically related at concept level, such as ``city,'' ``town,'' or ``place.''

In this paper, we present the idea of personalized word embeddings.
We explore differences in personalized word representations using a corpus of English Reddit posts that contains a large number of posts per author.
We use the embeddings to initialize a language model and show that personalization leads to better results than generic embeddings. 
One potential application of this work is personalized text generation for auto-completion to speed up text entry.
Another application is dialog systems that follow the speaking style of certain professionals (e.g., counselors, advisors).
Finally, personalized word representations could particularly help users with atypical writing styles that are not currently well served by models trained to suit the majority.

\section{Related Work}

Prior work has considered \emph{user embeddings}, where one vector is learned for each user in the data (we learn a set of vectors per user, one for each word in the vocabulary).
User embeddings have been used for dialog generation~\cite{li2016persona},
query auto-completion~\cite{jaech-ostendorf-2018-personalized}, authorship attribution~\cite{ebrahimi2016personalized}, and sarcasm detection~\cite{kolchinski-potts-2018-representing}.
\newcite{amer2016toward} learn a set of embeddings from the books that a user adds to their profile.
Some approaches also use network information \cite{zeng2017socialized,huang2016enriching}.

Personalization has been studied for marketing, webpage layout, recommendations, query completion, and dialog~\cite{eirinaki2003web,das2007google}.
Our prior work \cite{Welch19LearningFromPersonal,Welch19LookWhosTalking} explored predicting response time, common messages, and author relationships from personal conversation data. \newcite{zhang2018personalizing} conditioned dialog systems on artificially constructed personas and \newcite{madotto-etal-2019-personalizing} used meta-learning to improve this process. Goal-oriented dialog has used demographics (i.e., age, gender) to condition system response generation, showing that this relatively coarse grained personalization improves system performance \cite{joshi2017personalization}.
 
\section{Personalized Word Embeddings}\label{sec:generating_embeddings}

\paragraph{Definition.}
Personalized word embeddings are vector representations of words derived from the text produced by a single author.
We use the text produced by a Reddit user $s$ in their posts $C_s$ to create their word embeddings.
We apply the method described below to this set and produce an embedding matrix, $C_s \mapsto W_s^{|V|\times k}$, where $V$ is the vocabulary and $k$ represents the embedding dimension.

\paragraph{Joint Learning of Personal and Generic Word Embeddings.}
We jointly learn a generic embedding matrix and an embedding matrix for each author, inspired by \newcite{bamman2014distributed}. Each matrix $W \in \mathbb{R}^{|V|\times k}$ has a row for each vocabulary word and a k-dimensional vector for each embedding. The hidden layer is calculated as $h=w^\intercal W_{generic} + w^\intercal W_s$ where $w$ represents the one-hot encoding of a word and $s$ represents an author. This is a modified skip-gram architecture~\cite{mikolov2013efficient}, which sums two terms so that back-propagation updates the generic matrix and a author-specific matrix. It allows the generic matrix to benefit from all data while learning author-specific deviations in the same space.

\begin{table}
    \small
    \centering
    \bgroup
    \begin{tabular}{p{5mm} p{6.0cm} p{7.6cm}}
        \toprule
        User & Example Use & Nearest Neighbors \\
        \midrule
        A
        & doctors think this is bad for her \textbf{health} ...
        & preventative, insurance, reform, medical, education
        \\
        B
        & it is usually bad for your \textbf{health} ...
        & professional, mental, conduct, experiences, online
        \\
        All & N/A & medical, preventative, insurance, safety, healthcare
        \\
        \bottomrule
    \end{tabular}
    \egroup
    \caption{\label{tab:examples_change}
    Nearest neighbors of ``health'' for two personalized embedding spaces and the generic space.}
\end{table}

\paragraph{Dataset.}
We use data for the 100 most active users\footnote{
We excluded users that appear on a 
\url{https://www.reddit.com/r/autowikibot/wiki/redditbots}{public list of bots}
or who appear to be automated based on manual inspection.
}
in a corpus collected from Reddit.
These users have from 49k to 249k posts, with 73k on average.
Posts contain 29 tokens on average and come from 3.6k subreddits.
The largest fraction (18.6\%) belong to the subreddit \textit{AskReddit}; the next two largest are \textit{blog} and \textit{politics} with 4.8\% and 4.7\% of the posts respectively.
\smallskip

We use the set of messages from all 100 authors to generate embeddings for all words that occur at least five times (across all users). This yields a vocabulary of 177k words. We learn 100-dimensional embeddings with an initial learning rate of 0.025 and a window size of five, using L2 regularization due to the increased number of parameters (tuned in preliminary experiments).
Using this method, we learn 101 embeddings for each word -- a generic representation, and a separate representation for each user.

Reddit users have been found to be primarily male, young adults (under 30), located in the USA and primarily identify as christian or atheist~\cite{emnlp20compositional}. It is possible that results presented in this paper do not generalize to populations that differ significantly from the population of Reddit users. Future work may consider isolating the effects of topics and style by modeling subreddits for comparison though in this work we consider a personalized embedding as a representation that may capture both.

\input{word_type_dists.tex}

\section{Differences across Individual Word Representations and Usages}\label{sec:personal}

Individuals use the same word in different ways in different contexts. Examining these differences can give insight into individual topic and style preferences, or their word associations. 
To illustrate these differences, in Table \ref{tab:examples_change}, we show different ways that two users in our dataset use the word ``health.'' Although these words may be used in similar contexts, the meaning of, and topics associated with these words is often different for each user, which affects the words we would expect to come after it. These preferences are reflected in the top neighbors for the word ``health'' for each user.

To gain a deeper understanding of these differences, we analyze personal and generic embeddings for specific word groups based on the Linguistic Inquiry and Word Count (LIWC) lexicon \cite{pennebaker2001linguistic} and part-of-speech tags. This analysis can help us understand what types of words tend to have different representations across users and are therefore more personal in nature.
We part-of-speech tag the messages with the Stanford CoreNLP tagger~\cite{toutanova2003feature,manning2014stanford}.
For each word in the vocabulary, we assign a tag if the tagger gives the same tag at least 95\% of the time, otherwise the word is ignored. LIWC categories are looked up in the lexicon that contains words and word stems and a word may have multiple categories, in which case it counts toward each.

We look at the proportion of word types in the 5k most dissimilar words for each user.
We define word similarity as the cosine distance between a word's generic embedding and its author-specific embedding. Note that theses are unique words and that the x-axis is the percentage of the way through the 5k dissimilar words, with the most dissimilar at 100\%.
We break this into subsets and look at how the distribution changes as we approach the most dissimilar words. A visualization is provided in Figure \ref{fig:word_type_dists}. We find the set of most dissimilar words includes more function words, words relating to space and time (Relativ), cognitive processes (CogProc), and social words, as well as more adjectives and nouns. This suggests that these types of words may tend to have more personal usage than other types.
These results are consistent with prior work that has found function words are effective for recognizing style and measuring style similarity \cite{gonzales2010language} and for authorship attribution~\cite{mosteller1963inference,gamon2004linguistic,argamon2003style}. 

\section{Language Modeling}\label{sec:language_model}

Our first test case for personalized word embeddings is language modeling.
We train language models on our data with our embeddings as input. 
We use AWD-LSTM language model from previous work as our baseline~\cite{merity2018regularizing,merity2018analysis}.
It is an autoregressive language model that has state-of-the-art results by combining  regularization techniques and has been widely used.
Although more recent models can achieve better perplexities on standard benchmarks \cite{melis2019mogrifier,dai-etal-2019-transformer}, we find that not all models have code available or that they take far more time to run than \newcite{merity2018regularizing}'s model.
We use the same hidden layer sizes and drop-out rates as in their original experiments, but untie the weights of the encoder and decoder as that gave better performance in preliminary experiments.

To use our personalized embeddings, we modify the architecture to take as input the concatenation of the personalized user-specific embedding and the generic embedding for each word.
The same embedding dropout mask is applied to both word embeddings.
The embeddings are trained on all available user data, but the more computationally expensive language models are not.
We use a subsample of our dataset with 1,000 posts for each user and an 80/10/10 split for training, validation, and testing.
The same splits are used for generic and personalized models, varying only the embedding layer.

To measure the ability of our models to predict the next word, we use two metrics: (1) mean reciprocal rank (MRR), calculated as one divided by the rank of the correct word choice in the descending list of next word probabilities and averaged over all instances, and (2) perplexity.

\vskip 0.1in
\noindent \textbf{Single User Embeddings.}
We also consider an approach in which just one vector is learned for each user (rather than one for each user-word pair). This is an approach widely used in previous work~\cite{kolchinski-potts-2018-representing,li2016persona}. This user vector is concatenated to the generic word embedding.

The results in Table \ref{tab:results_all} suggest that using the combined personalized and generic embeddings improves performance significantly over single vector user representations and over generic embeddings. We note that the number of parameters of the LSTM input is not the same when comparing the baseline \newcite{merity2018regularizing} model to the other cases. We ran an additional experiment doubling the size of the embeddings for the baseline and found that the perplexity improved to 64.21, although the personalized embeddings still significantly outperformed this baseline.

We can also analyze the accuracy when predicting words belonging to particular parts of speech and LIWC categories. 
Our intuition is that a model that uses personalized word embeddings would be better at predicting words belonging to the four LIWC categories whose words are most distant from the generic space. Tables \ref{tab:pos_results} and \ref{tab:results_liwc} show that for almost all categories, the personalized word embeddings lead to the best performance, although for relativity words, single user vectors performs slightly better.

\section{Authorship Attribution}\label{sec:authorship}

We also use a language model trained with personalized word embeddings to perform the task of authorship attribution.\footnote{Note that we do not consider datasets such as~\newcite{pan} because they do not provide the volume of data needed for our approach and our goal is to compare generic and personalized embeddings, not to set a new state-of-the-art.}
We build a language model for each author using a sample of 10k posts for training and 1k for validation. We then hold out another sample of 1k posts to use for authorship attribution. The language models for all authors are separately run on the held out set, and the model with the lowest perplexity is then chosen as the assigned author. 
Table~\ref{tab:results_all} shows there is a statistically significant improvement for our personalized embeddings method.
This is a difficult task with 100 classes, so the accuracy is low, but the MRR suggests that the correct author is usually in the top 3 model choices.

\begin{table}
    \centering
    \footnotesize
    \begin{tabular}{l c cc c cc}
        \toprule
              & & \multicolumn{2}{c}{LM} & & \multicolumn{2}{c}{Attribution} \\
        Model & & MRR & PPL & & MRR & Accuracy \\ \midrule
        \newcite{merity2018regularizing} & & 0.364 & 65.53 & & 0.452 & 34.8\% \\
        Single User Vectors & & 0.361 & 66.70 & & 0.450 & 34.9\% \\
        Personalized Embeddings & & \textbf{0.371} & \textbf{62.43} & & \textbf{0.462} & \textbf{36.1\%} \\
        \bottomrule
    \end{tabular}
    \caption{
    \label{tab:results_all}
    Results for Language Modeling (LM) and Authorship Attribution.
    Personalized word embeddings significantly improve performance (permutation test, $p<0.0001$).
    }
\end{table}

\begin{table*}
    \footnotesize
    \centering
    \begin{minipage}[]{\linewidth}
    \centering
    \setlength{\tabcolsep}{8pt}
    \begin{tabular}{lccccccccc}
        \toprule
        Model & DT & IN & JJ & NN & PR & RB & VB & PUNCT & OTHER \\ \midrule
        \newcite{merity2018regularizing} & 19.1 & 30.9 & 703.6 & 632.5 & 23.4 & 146.6 & 65.4 & 10.9 & 72.6 \\
        Single User Vectors & 19.4 & 34.2 & 708.2 & 621.4 & 23.7 & 148.7 & 65.8 & 11.1 & 73.9 \\
        Personalized Word Embeddings & \textbf{18.9} & \textbf{30.7} & \textbf{681.2} & \textbf{597.7} & \textbf{22.6} & \textbf{143.7} & \textbf{62.1} & \textbf{10.2} & \textbf{70.2} \\
        \bottomrule
    \end{tabular}
    \subcaption{
    \label{tab:pos_results}
    Perplexity results broken down by POS tag (OTHER includes all other tags).}
    \end{minipage}
    \vspace{2mm}
    
    \begin{minipage}[]{\linewidth}
    \footnotesize
    \centering
    \begin{tabular}{lccccccccc}
        \toprule
        Model & Affect & Bio & CogProc & Drives & Funct & Inform & Percept & Relativ & Social \\ \midrule
        \newcite{merity2018regularizing} & 88.3 & 93.9 & 73.4 & 69.8 & 68.8 & \textbf{35.3} & 95.9 & 28.7 & 48.7 \\
        Single User Vectors & 85.6 & 95.7 & 75.8 & 75.0 & 71.2 & 36.0 & 90.7 & \textbf{28.1} & 48.9 \\
        Personalized Word Embeddings & \textbf{82.7} & \textbf{92.4} & \textbf{73.3} & \textbf{70.7} & \textbf{67.6} & \textbf{35.3} & \textbf{85.4} & 28.5 & \textbf{46.7} \\
        \bottomrule
    \end{tabular}
    \subcaption{
    \label{tab:results_liwc}
    Perplexity results broken down by high-level LIWC category.}
    \end{minipage}
    \caption{
    Perplexity results broken down by the type of target word, with the best result in bold.}
\end{table*}

\section{Limitations and Ethical Considerations}

In applying our method to text prediction systems, users may experience unintended negative effects. For instance, embeddings may become unintentionally biased toward language that becomes inappropriate when later suggested in another context. Additionally, users who are learning a language may bias embeddings toward improper language use, reinforcing errors and making it more difficult for the user to learn the language. It may be appropriate to use our embeddings if users consent and are made aware of the possible consequences of doing so.

It is possible that our method could be used for authorship attribution and surveillance of individuals online~\cite{stamatatos2009survey}. Such an application risks potential discrimination, coercion, and threats to intellectual freedom~\cite{richards2012dangers}. Personalized language models could also be used to develop a tool that tells the user who their writing most resembles, or if their writing resembles their past writing, with the objective of obfuscating the author's identity~\cite{potthast2018overview}. A tool like this could also be used maliciously to impersonate a particular author. Although we believe the difficulty of this task currently makes these minor risks, we advocate against the use of our methods for these tasks.

Our method requires more computation and memory than the baseline method we compared to. The additional computation is relatively small, as learning the embeddings takes around 3 hours using 30 threads on a machine with 16 Intel Xeon Silver 4108 CPUs. The memory required to store embeddings for $N$ users is $N+1$ times the amount of storage required by a generic matrix only.

\section{Conclusion}

In this paper, we explored personalized word embeddings. Using a large corpus of Reddit posts, we generated personalized word embeddings for 100 individuals, and performed analyses of the differences between personalized and generic embeddings for specific groups of words. We showed that using personalized word embeddings to initialize a language model improves perplexity over a model that uses generic word embeddings, or a model that only learns single vectors for each user as has been frequently done in previous work. Further, we showed that the embeddings can be used to improve performance on authorship attribution. We cannot release the data due to licensing restrictions but our code is available online with instructions for how to obtain and process the data in order to support future work on personalization.\footnote{\url{https://lit.eecs.umich.edu}}

\section*{Acknowledgements}

We would like to thank the anonymous reviewers for their helpful suggestions.
This material is based in part on work supported by IBM (Sapphire Project), DARPA (grant \#D19AP00079), Bloomberg (Data Science Research Grant), the NSF (grant \#1815291), and the John Templeton Foundation (grant \#61156). Any opinions, findings, conclusions, or recommendations in this material are those of the authors and do not necessarily reflect the views of IBM, DARPA, Bloomberg, the NSF, or the John Templeton Foundation. 

\bibliography{main}
\bibliographystyle{coling}

\end{document}